%% file: main.tex
\documentclass[twoside,leqno,twocolumn]{article}
\usepackage[letterpaper]{geometry}

\usepackage{ltexpprt}
\usepackage[utf8]{inputenc}
\usepackage{amssymb}
\usepackage{amsmath}
\usepackage{amsfonts}
\usepackage{listings}
\usepackage{caption}
\usepackage{bm}

\usepackage{algorithm}
\usepackage{algorithmic}
\usepackage{wrapfig}

\usepackage{xspace}
\usepackage{todonotes}
\usepackage{mathtools}
\usepackage{mathrsfs}
\usepackage{xfrac}
\usepackage{booktabs}
\usepackage{multirow}

\usepackage{array,multirow,colortbl,graphicx,subcaption}
\usepackage{caption}
\usepackage{adjustbox}
\usepackage{url}
\newcommand{\aaa}{autoencoder}
\newcommand{\algo}{BAE\xspace}
\newcommand{\prob}{${\rm I\!P}^{(i)}$\xspace}
\NewDocumentCommand{\rot}{O{90} O{1em} m}{\makebox[#2][r]{\rotatebox{#1}{#3}}}%

\graphicspath{{img/}}
\begin{document}

\title{\Large  Unsupervised Boosting-based Autoencoder Ensembles for Outlier Detection}
\date{}

\author{
Hamed Sarvari\thanks{hsarvari@gmu.edu, George Mason University, Fairfax, VA}
\and Carlotta Domeniconi\thanks{cdomenic@gmu.edu, George Mason University, Fairfax, VA}
\and Bardh Prenkaj\thanks{prenkaj@di.uniroma1.it, Sapienza University of Rome, Italy}
\and Giovanni Stilo\thanks{giovanni.stilo@univaq.it, University of L'Aquila, Italy}
}

\maketitle








\input{abstract.tex}

\input{sec_intro-hamed.tex}
\input{sec_related-work.tex}

\input{sec_methodology.tex}
\input{sec_experiments.tex}

\input{sec_conclusion.tex}
\bibliography{References}
\bibliographystyle{ieeetr}

\end{document}

%% file: abstract.tex
\begin{abstract} \small\baselineskip=9pt 
Autoencoders, as a dimensionality reduction technique, have been recently applied to outlier detection. However, neural networks are known to be vulnerable to overfitting, and therefore have limited potential in the unsupervised outlier detection setting. Current approaches to ensemble-based autoencoders do not generate a sufficient level of diversity  to avoid the overfitting issue. 
To overcome the aforementioned limitations we develop a Boosting-based Autoencoder Ensemble approach (in short, \algo). \algo is an unsupervised ensemble method that, similarly to the boosting approach, builds an adaptive cascade of autoencoders to achieve improved and  robust results.
\algo trains the autoencoder components sequentially by performing a weighted sampling of the data, aimed at reducing the amount of outliers used during training, and at injecting  diversity in the ensemble.
We perform extensive experiments and show that the proposed methodology outperforms state-of-the-art approaches under a variety of conditions.


\end{abstract}

%% file: sec_intro-hamed.tex
\section{Introduction}


Outlier (or anomaly) detection is the process of automatically identifying irregularity  in the data. An outlier is a piece of data or observation that deviates drastically from the given norm or average of the dataset. This is a widely accepted definition of an anomaly. Nevertheless, outlier detection, being  intrinsically unsupervised, is an under-specified, and thus {\textit{ill-posed}}, problem. This makes the task particularly challenging. 

Several anomaly detection techniques have been introduced in the literature, e.g. distance-based~\cite{knox1998algorithms, ramaswamy2000, angiulli2002fast, ghoting2008fast, fan2006nonparametric} , density-based~\cite{lof, jin2001mining, tang2002enhancing, papadimitriou2003loci}, and subspace-based methods~\cite{aggarwal2000finding,keller2012hics,lazarevic2005feature}. 
Neural networks, and specifically autoencoders, have also been used for outlier detection \cite{williams2002comparative,hawkins2002outlier, chen2017outlier}. An autoencoder is a multi-layer symmetric neural network whose goal is to reconstruct the data provided in input.  To achieve this goal, an autoencoder learns a new reduced representation of the input data that minimizes the reconstruction error. When using an autoencoder for outlier detection, the reconstruction error indicates the level of outlierness of the corresponding input.

As also discussed in \cite{chen2017outlier}, while deep neural networks have shown great promise in recent years when trained on very large datasets, they are prone to overfitting when data is limited due to their model complexity. For this reason, autoencoders are not popular for outlier detection, where data availability is an issue.


The authors in \cite{chen2017outlier} have attempted to tackle  the aforementioned challenges by generating randomly connected autoencoders, instead of fully connected, thus reducing the number of parameters to be tuned for each model. As such, they obtain an ensemble of autoencoders, whose outlier scores are eventually combined to achieve the ultimate scoring values. The resulting approach is called RandNet.

Ensembles have the potential to address the ill-posed nature of outlier detection, and they have been deployed with success to boost performance in classification and clustering, and to some extent in outlier discovery as well \cite{chen2017outlier,Aggarwal13,Aggarwal15, schubert12,Nguyen10,rayana2015less, sarvari2019graph}. The aggregation step of ensembles can filter out spurious findings of individual learners, which are due to the specific learning bias being induced, and thus achieve a consensus that is more robust  than the individual components. To be effective, though, an ensemble must achieve a good trade-off between the accuracy and the diversity of its components.



In this paper we focus on both autoencoders and the ensemble methodology to design an effective approach to outlier detection, and propose a \textbf{B}oosting-based \textbf{A}utoencoder \textbf{E}nsemble method (\algo). Our analysis (see section~\ref{subsec:div}) shows that RandNet~\cite{chen2017outlier} may not achieve a good balance between diversity and accuracy, and thus is unable to fully  leverage the  potential of the ensemble. Unlike RandNet, which trains the autoencoders independently of one another, our approach uses fully connected autoencoders, and aims at achieving diversity  by performing an adaptive weighted sampling, similarly to boosting but for an unsupervised scenario. 
We train an ensemble of autoencoders in sequence, where the  data sampled to train a given component depend on the reconstruction errors obtained by the previous one; specifically,  the larger the error, the smaller the weight assigned to the corresponding data point is. As such, our adaptive weighted sampling progressively forces the autoencoders to focus on inliers, and thus to learn representations which  lead to large reconstruction errors for the outliers. This process facilitates the generation of  components with accurate outlier scores. We observe that, unlike standard supervised boosting, our  weighting scheme is not prone to overfitting because we proportionally  assign more weight to the inliers, which are the ``easy to learn" fraction of the data. 

Overall the advantage achieved by BAE is twofold. 
At one end, the progressive reduction of outliers enables the autoencoders to learn better representations of  ``normal" data, which also results in accurate outlier scores. At the other end, each autoencoder is exposed to a different set of outliers, thus promoting diversity among them. Our experimental results show that this dual effect indeed results in a good accuracy-diversity trade-off, and ultimately to a competitive or superior performance against state-of-the-art competitors.

The rest of this paper is organized as follows. Section~\ref{sec:rel-work} briefly discusses related work. In Section~\ref{sec:methodology}, we present our approach in detail. Sections ~\ref{sec:experiments} and ~\ref{sec: discussion}  present and discuss the empirical evaluation. Section~\ref{sec:conclusions} concludes the paper.



%% file: sec_related-work.tex

\section{Related work}\label{sec:rel-work}


Various outlier detection techniques have been proposed in the literature, ranging from distance-based~\cite{knox1998algorithms, ramaswamy2000, angiulli2002fast, ghoting2008fast, fan2006nonparametric}, density-based~\cite{lof, jin2001mining, tang2002enhancing, papadimitriou2003loci}, to subspace-based~\cite{aggarwal2000finding,keller2012hics,lazarevic2005feature} methods. A  survey of anomaly detection methods can be found in~\cite{chandola2007outlier}.
Neural networks, as a powerful learning tool, have also been used for outlier detection, and
autoencoders are the fundamental architecture being deployed. Hawkins~\cite{hawkins2002outlier} used autoencoders for the first time to address the anomaly detection task. 
Deep autoencoders, though, are known to be prone to over-fitting when limited data are available~\cite{chen2017outlier}.
A survey on different neural network based methods used for the anomaly discovery task can be found in~\cite{chalapathy2019deep}.

Autoencoders are also used in semi-supervised novelty detection scenarios, in which  access to a pure inlier class is assumed. In such techniques, an underlying model of regularity is constructed from the normal data, and future instances that do not conform to the behaviour of this model are flagged as anomalies. A comprehensive survey of these methods can be found in~\cite{kiran2018overview}. A  deep autoencoder, based on robust principal component analysis, is used in~\cite{zhou2017anomaly} to remove noise, and discover high quality features from the training data. 

The ensemble learning methodology has been used in the literature to make the outlier detection process more robust and reliable \cite{Aggarwal13,Aggarwal15, schubert12,Nguyen10,schubert12,rayana2015less, sarvari2019graph}. For example, HiCS~\cite{keller2012hics} is an ensemble strategy which finds high contrast subspace projections of the data, and aggregates outlier scores in each of those subspaces. HiCS shows its potential when outliers do not appear in the full space and are hidden in some subspace projections of the data.

Ensembles can dodge the tendency of autoencoders to overfit the data~\cite{chen2017outlier}. Only a few ensemble-based outlier detection methods using autoencoders have been proposed. RandNet~\cite{chen2017outlier} introduces the idea of using an ensemble of randomly connected autoencoders, where each component has a number of connections randomly dropped. The median outlier score of the data across the autoencoders is considered as the final  score.  

Our  experiments  show  that  RandNet  may  not achieve a good balance between diversity and accuracy, and  thus  is  unable  to  fully  leverage  the  potential  of the  ensemble. To better leverage autoencoders within the ensemble methodology, we propose a boosting-based approach ~\cite{freund1997decision, freund1996experiments} aimed at generating both diverse and accurate components.



%% file: sec_methodology.tex

\section{Methodology}\label{sec:methodology}

We build an ensemble inspired by the boosting algorithm approach \cite{freund1997decision} that uses \aaa s. 
In the boosting context, a weak learner is trained sequentially to build a stronger one. Likewise, we  train a sequence of \aaa s on different data sub-samples, thus building a boosted ensemble \algo for anomaly detection. 
At each step of the boosting cascade, we sample a new dataset from the original one by exploiting only the reconstruction errors, remaining fully unsupervised unlike the original boosting strategy.
Moreover, using different subsequent samples lowers the overall learning bias and enhances the variance. 
Finally, the components are combined into a weighted sum that represents the final output of the boosted model. 

From now on, we refer to each step of the boosting cascade as an iteration $i \in [0,m)$ of \algo producing a component of the final ensemble.
The number of iterations, as the size of the ensemble $m$, is a parameter of \algo.
Let AE$^{(i)}$ be the autoencoder trained on the sample $\mathbf{X}^{(i)}$ obtained in the $i$-th iteration of the boosting cascade. 

When detecting anomalies, it is common to use the reconstruction error of an \aaa \xspace as indicator of the outlierness level of each data point. 
The rationale behind this approach is that the autoencoders represent an approximated identity function. Therefore, the applied model is compelled to learn the common characteristics of the data. The learnt data distribution does not include the characterization aspects of outlier instances thus generating higher reconstruction errors.

As stated above, in \algo the boosting approach is achieved by guiding the learning phase changing the set of data points of each iteration. In detail, at each iteration $i \in [0,m)$, instances of the new dataset $\mathbf{X}^{(i)}$ are sampled with replacement from the original one $\mathbf{X}^{(0)}$ according to a probability function \prob.
\begin{figure*}[t]
    \centering
    \begin{subfigure}[b]{0.5\textwidth}
        \centering
        \includegraphics[width=1\linewidth]{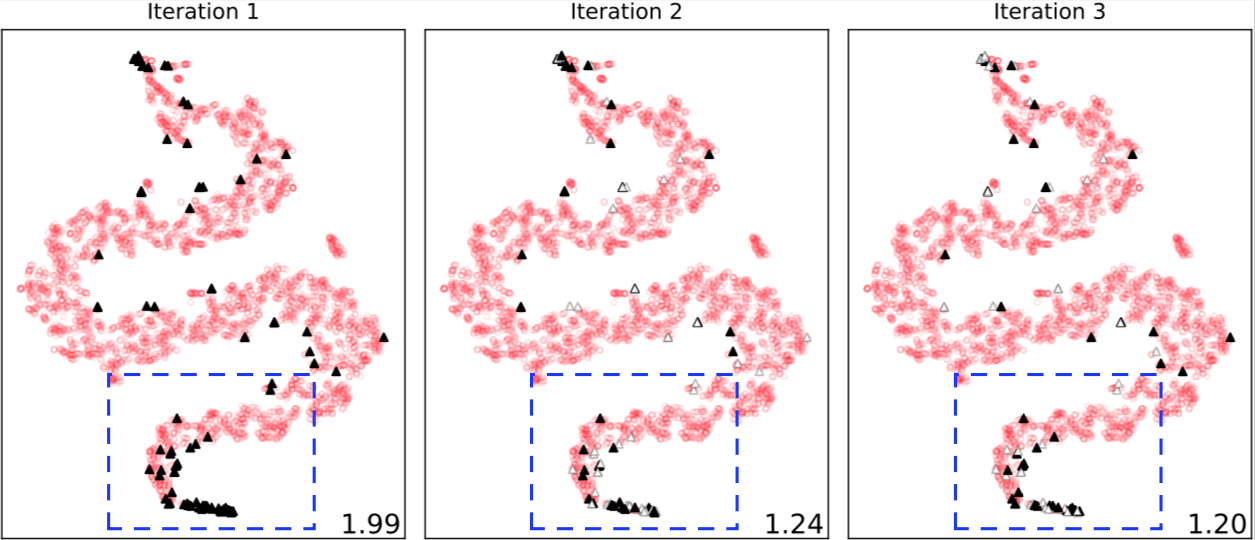}
    \end{subfigure}%
    ~ 
    \begin{subfigure}[b]{0.5\textwidth}
        \centering
        \includegraphics[width=1\linewidth]{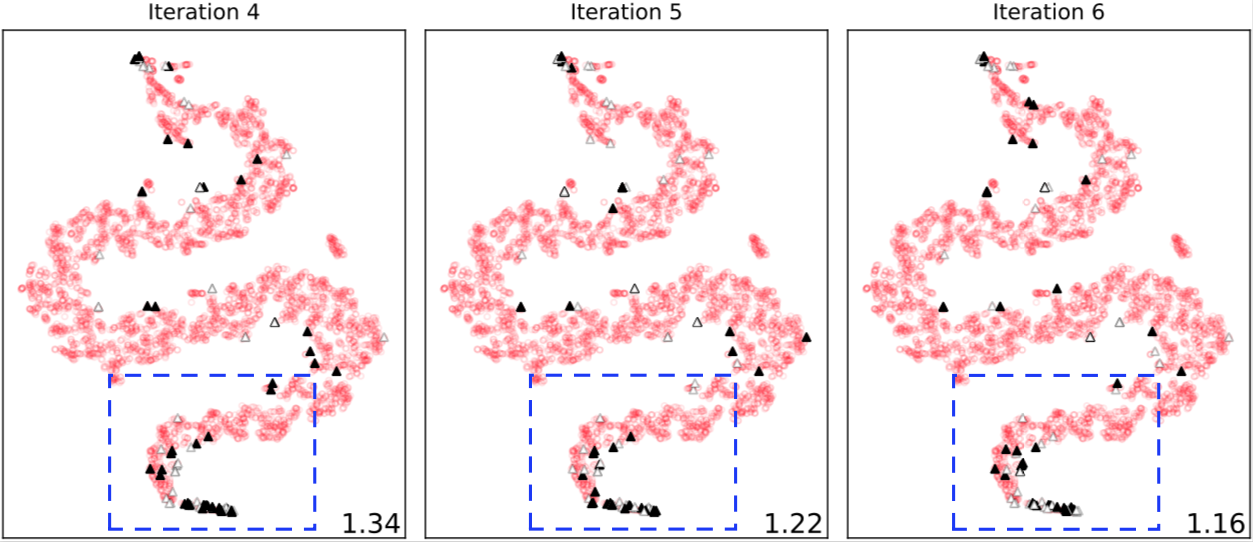}
    \end{subfigure}
    \caption{The ratio of outliers in the different samples $\mathbf{X}^{(i)}$ for PageBlocks. \label{fig:dragon}}
\end{figure*}

Following the autoencoder approach described above, we build a probability function \prob based on the reconstruction error of the original dataset $\mathbf{X}^{(0)}$ obtained from the \aaa \xspace of the previous iteration $i-1$. Essentially, we want to have a sampled dataset $\mathbf{X}^{(i)}$ that has fewer outliers w.r.t. $\mathbf{X}^{(i-1)}$.
To achieve this,  we define a distribution that assigns weights to data which are inversely proportional to the corresponding reconstruction error. Therefore, the function \prob gives higher probability to those data points that have a lower reconstruction error (inliers) in the following way: 
$${\rm I\!P}^{(i+1)}_x=\frac{1/e^{(i)}_{x}}{\sum{1/e^{(i)}_{x}}} $$
where ${\rm I\!P}^{(i+1)}_x$ is the probability of $x$ to be selected while sampling the new dataset, and $e^{(i)}_x$ is defined as:
$$e^{(i)}_x =\Big(||x - AE^{(i)}(x)||_2\Big)^2$$
$e^{(i)}_x$ is the reconstruction error of data point $x \in \mathbf{X}^{(0)}$ obtained using the \aaa \xspace AE$^{(i)}$ of iteration $i$. We observe that,  building a new probability function ${\rm I\!P}^{(i+1)}$ at each iteration $i$, induces the data sampling to create a brand new dataset, and this produces a different component of the ensemble after the training phase of the corresponding \aaa.


According to \prob, $\mathbf{X}^{(i)}$ contains more inliers than $\mathbf{X}^{(i-1)}$. Therefore, our ensemble method specializes in the inlier instances, discriminating the outlier ones with each iteration. Figure \ref{fig:dragon} demonstrates this observation. We exploit BAE with 5 layers for this example. The number in the bottom right corner of the plots is the percentage of remaining outliers in $\mathbf{X}^{(i)}$ w.r.t. the overall number of instances in $\mathbf{X}^{(0)}$. Figure \ref{fig:dragon} shows the overall  declining trend of this ratio through the iterations. We represent with red-contoured circles\footnote{We use two colors (b\&w) to highlight the selection of the outliers. The inliers are represented without a color variation.} the inliers of $\mathbf{X}^{(0)}$. Filled triangles denote the selected outliers for sample $\mathbf{X}^{(i)}$ and unfilled ones are those discarded from $\mathbf{X}^{(i)}$. Finally, we observe that the tail of the distribution on the bottom part of the plots tends to thin out. In other words, after each iteration the tail contains more white triangles than black ones.






The characteristic of an ensemble method lies also into combining the results of the components, denoted as consensus function. In our case, the components are inlier-specialized and are better suited for detecting outliers.
Following the above observation, the component that has smaller reconstruction errors has more impact on the final anomaly score. Thus, we build a weighted consensus function that assigns weights to each component based on the sum of reconstruction errors that the \aaa s generate on their corresponding sample. Notice that the first component is based on the original dataset and not on a sampled one. 
To be consistent with the boosting strategy, our method uses several autoencoders to improve the performance achieved by a single autoencoder (the first iteration). We decide to discard the reconstruction errors produced by the first iteration AE$^{(0)}$. Moreover, empirical evaluation (see Section \ref{sec:experiments}) confirms our choice because it shows that a single autoencoder (the first iteration) performs worst than \algo.
Therefore, the weight $w^{(i)}$ of the $i$-th iteration is formally defined as follows:

$$w^{(i)}=\frac{\sfrac{1}{\sum\limits_{x \in \mathbf{X}^i}{e^{(i)}_x}}}{\sum\limits_{i=1}^{m-1}\left({\sfrac{1}{\sum\limits_{x \in \mathbf{X}^{(i)}}{e^{(i)}_x}}}\right)} $$

Hence, the final outlier score $\bar e_x$ assigned to each datum $x \in \mathbf{X}^{(0)}$ is a weighted sum of the reconstruction errors of $x$ in the sequence of the last $m-1$ components:
$$\bar e_x= \sum \limits_{i=1}^{m-1}{w^{(i)}e_x^{(i)}}$$
 

We summarize the aforementioned insights in the pseudo-code\footnote{\algo is available at \url{https://gitlab.com/bardhp95/bae}} presented in Algorithm \ref{algo:sambae}. The algorithm is divided into two phases: boosting-based training and consensus phase.

In the {\em boosting-based training phase}, we train our \aaa \xspace on the $i$-th view $\mathbf{X}^{(i)}$ of the original dataset (line 3). Then, we proceed on specializing by sampling the new dataset with the instances whose reconstruction error is lower. Hence, for each instance $x \in \mathbf{X}^{(i)}$ we calculate the probabilistic function ${\rm I\!P}^{(i+1)}$, based on the reconstruction errors of the instances (lines 4-7), and compose the new dataset $\mathbf{X}^{(i+1)}$ according to it (line 8). The function sample($\mathbf{X}, {\rm I\!P}$) returns a new dataset $\tilde{\mathbf{X}}$ based on the  probability distribution ${\rm I\!P}$.  We repeat the training phase for all the $m$ components of the ensemble before continuing with the consensus phase.

We use the weighting function of the {\em consensus phase} to distribute the authority scores to each component of the ensemble. We calculate it for the last $m-1$ components on the sampled data $\mathbf{X}^{(i)}$ (lines 11-13). The components with higher weights (i.e. lower reconstruction error) contribute more to the final reconstruction error of the data. Weights of each component and the probability score of each instance  are normalized (notice the denominators in line 6 and 12). 

Finally, we merge the result as the weighted sum of the reconstruction errors of the data points producing the outlier scores (lines 14-15).
\begin{algorithm}[H]
\caption{\algo to calculate scoring vector $\bar e$}\label{algo:sambae}
\begin{algorithmic}[1] 
\REQUIRE $\mathbf{X}^{(0)}, m$
\STATE \textit{--- Boosting-based training phase ---}
\FOR{$i=0$ to $m-1$}
    \STATE Train AE$^{(i)}$ on $\mathbf{X}^{(i)}$
    \FOR{$x \in \mathbf{X}^{(i)}$}
        \STATE $e^{(i)}_x \leftarrow$ $\Big(||x -$AE$^{(i)}(x)||_2\Big)^2$
        \STATE ${\rm I\!P}^{(i+1)}_x \leftarrow \frac{1/e^{(i)}_{x}}{\sum{1/e^{(i)}_{x}}}$
    \ENDFOR
    \STATE $\mathbf{X}^{(i+1)} \leftarrow$ sample($\mathbf{X}^{(0)}$, ${\rm I\!P}^{(i+1)}$)
\ENDFOR
\STATE \textit{--- Consensus phase ---}
\FOR{$i=1$ to $m-1$}
    \STATE $w^{(i)} \leftarrow \frac{\sfrac{1}{\sum\limits_{x \in \mathbf{X}^{(i)}}{e^{(i)}_x}}}{\sum\limits_{i=1}^{m-1}\left({\sfrac{1}{\sum\limits_{x \in \mathbf{X}^{(i)}}{e^{(i)}_x}}}\right)}$
\ENDFOR
\FOR{$x \in \mathbf{X}^{(0)}$}
    \STATE$\bar e_x \leftarrow \sum \limits_{i=1}^{m-1}{w^{(i)}e_x^{(i)}}$
\ENDFOR
\RETURN $\bar e$
\end{algorithmic}
\end{algorithm}


\subsection*{Neural Network Details:}\label{sub:meth_architecture} 
As base architecture of our ensemble method we use a fully connected \aaa, where all neurons of one layer are connected to all the neurons of the successive layer. An autoencoder is a special network with a symmetric design where the number of input and output neurons is equal to the data dimensionality denoted as $d_0$. 

Then, we need to decide the number of neurons in each internal layer, the activation functions, and the number of layers (i.e. depth denoted with $l$). We choose the number of hidden neurons following the strategy proposed in \cite{chen2017outlier}. Hence, we model all the hidden layers of our encoder in the following way: 
$$d_{h} =  \lfloor{\alpha \cdot d_{h-1}}\rfloor $$ 
where $h$ is the $h$-th layer of the encoder. The decoder's hidden layer sizes are symmetrical to the encoder ones. If $\lfloor{\alpha \cdot d_{h-1}}\rfloor$ is less than 3 for layer $h$, then the number of neurons in that layer is set to 3. This avoids the excessive compression in the hidden (including the bottleneck) layers such that the \aaa \xspace has difficulty in properly reconstructing the input.

We choose a Sigmoid activation function for the first hidden layer and the output layer. Meanwhile, for all other layers we use the Rectified Linear (ReLU) activation function \cite{nair2010rectified}. The reason for selecting two different activation functions is twofold. First, because the Sigmoid function is prone to the vanishing gradient descent problem, we use the ReLU unit to mitigate this phenomenon. Second, ReLU suffers from the "dying ReLU" problem\footnote{\url{http://cs231n.github.io/neural-networks-1/}}, as analyzed also in \cite{Lu2019DyingRA}, which causes neurons to be stale when dealing with large gradients. 
This motivates the usage of Sigmoid functions in the first and last layers.

\begin{figure*}[ht]
    \centering
	\includegraphics[width=\textwidth]{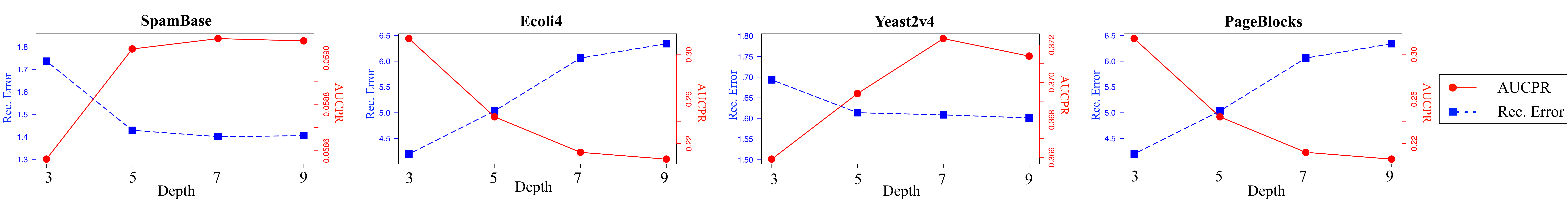}
\vspace{-1.8em}
\caption{Relationship between average reconstruction error of training samples across all ensemble components and performance (AUCPR) for different depths of the ensemble autoencoders.}
\label{fig:auc_vs_err}
\end{figure*}

To decide the number of layers $l$, we introduce a methodology to optimize the depth of the autoendoders for each dataset. We execute \algo for each dataset for several depth values $\mathcal{L}$, 
and following the observed negative correlation between the reconstruction errors of the ensembles and the AUCPR scores (see Figure~\ref{fig:auc_vs_err} and subsection \ref{subsection:ens-depth} for details), we build an optimization strategy that chooses the depth that resulted in the minimum reconstruction error.
Formally, we choose the $l$ that minimizes the following optimization function:
$$argmin\underset{l \in \mathcal{L}}{}
\Bigg\{\frac{1}{m-1}\sum\limits_{i=1}^{m-1}\sum\limits_{x \in \mathbf{X}^{(i)}} \Big(||x-AE^{(i)}_l(x)||_2\Big)^2\Bigg\}$$

Where $AE^{(i)}_l$ is the \aaa \xspace of depth $l$ trained at iteration $i$.

%


%% file: sec_experiments.tex

\section{Experiments}\label{sec:experiments}
\input{tbl/tbl_ds_sizes.tex}

\textbf{Methods:} 
To assess the effectiveness of our approach, we compare our performance against a number of baseline techniques: four outlier detection algorithms, namely
LOF~\cite{breunig2000lof}, a single autoencoder with nine layers (in short, SAE9), one-class SVM~\cite{scholkopf2001estimating}, and Hawkins~\cite{hawkins2002outlier}; and two ensemble techniques, i.e.
 HiCS~\cite{keller2012hics} and
RandNet~\cite{chen2017outlier}.


LOF is a well-known density-based outlier detection method. For each run of LOF, the value of the MinPts parameter is randomly chosen from the set \{3, 5, 7, 9\}.
SAE9 is a single autoencoder with nine layers. The network structure of SAE9 is the same as the components of \algo, as discussed in Section~\ref{sec:methodology}.
One-class SVM (OCSVM) estimates the support of a distribution. We used an RBF kernel with length-scale parameter set to 1. The soft margin parameter is 0.5, and the tolerance for the stopping criterion is 0.001. 
As the authors suggest~\cite{hawkins2002outlier}, for Hawkins we set the number of layers to 5.
To set the parameters of HiCS, we follow the guidelines in \cite{keller2012hics}. We use 50 Monte-Carlo iterations, $\alpha=0.05$, and a cutoff of 500 for the number of candidates in the subspaces. As base outlier detector algorithm we use LOF with 10 neighbors. For RandNet, as the authors suggest, 
we use 7 layers, and we set the structure parameter  $\alpha=0.5$. We set the growth ratio equal to 1 and we use 20 different ensemble components.



\textbf{Performance evaluation:} We use the area under the precision-recall curve (AUCPR) - often called average precision - to quantify the performance of each method. This measure is known to be more effective than the area under the ROC curve (AUCROC) when dealing with imbalanced datasets \cite{saito2015precision}.  

\textbf{Setting of BAE parameters:}\label{sec:exp_parameters}
We set the ensemble size to 20. We tested the sensitivity of BAE under different ensemble sizes, and observed that the performance is stable for sizes beyond 20. We train each \aaa \xspace AE$^{(i)}$ for 50 epochs. We stop earlier if the reconstruction errors converge, that is when the difference between the average reconstruction error at the end of two consecutive epochs is negligible (lower than $10^{-4}$).
We set the shrinking parameter $\alpha=0.5$ (see subsection \ref{sub:meth_architecture}). We use ADAM~\cite{kingma2014adam} as the optimizer of each AE$^{(i)}$ with learning rate $lr=10^{-3}$ and weight decay of $10^{-5}$.
We train \algo multiple times on various depths. Hence, for each dataset, we select the optimal depth on $\mathcal{L} = \{3,5,7,9\}$ following the strategy discussed in Section~\ref{sec:methodology}.

\textbf{Datasets:}
Data designed for classification are often used to assess the feasibility of outlier detection approaches. Similarly, here we use labeled data from the UCI Machine Learning Repository\footnote{\url{https://archive.ics.uci.edu/ml/datasets.php}} and adapt them to the outlier detection scenario. Instances from the majority class, or from multiple large classes, provide the inliers, and instances from the minority classes (or down-sampled classes) constitute the outliers. 

\input{Results-table.tex}

For ALOI, KDDCup99, Lymphography, Shuttle, SpamBase, WDBC, Wilt, and WPBC we follow the conversion process used in \cite{campos2016evaluation}\footnote{Data available at: \url{https://www.dbs.ifi.lmu.de/research/outlier-evaluation/DAMI/}}. Ecoli4, Pima, Segment0, Yeast2v4, and Yeast05679v4 were generated as described in \cite{alcala2011keel,alcala2009keel}\footnote{\url{https://sci2s.ugr.es/keel/imbalanced.php}}. In SatImage, the points of the majority class are used as inliers, and $1\%$ of the rest of the data is sampled to obtain the outliers. We also test the approaches on MNIST, a large image dataset of handwritten digits.
For MNIST, we generate a dataset for each digit, where the images of the chosen digit are the inliers. The outliers are obtained by uniformly  sampling 1\% of the other digits\footnote{Used data available at: \url{https://doi.org/10.6084/m9.figshare.9954986.v1}. New version with further information at: \url{https://doi.org/10.6084/m9.figshare.9954986.v2}}. The image of a digit is represented as a vector with $784$ dimensions.

We found that HiCS has difficulties with the larger data (i.e. KDDCup99 and MNIST) which induce its execution to not converge\footnote{The ELKI project implementation of HiCS produces a warning message and the execution does not reach a converging state.}. We have solved this problem by exploiting dimensionality reduction techniques for the aforementioned datasets and let HiCS successfully run. We use TSNE ~\cite{maaten2008visualizing} with three components for the reduction of KDDCup99, and PCA ~\cite{wold1987principal} with ten components for the reduction of the MNIST datasets. We then use the reduced versions of KDDCup99 and MNIST for HiCS only.

Finally, Table \ref{tab:datasets} reports a summary of the statistics of the datasets used in our experiments.

\textbf{Results:}
Table~\ref{tab:baselines} shows the performance achieved by all the tested methods. We run each technique 30 times, and report the average AUCPR values. Statistical significance  is evaluated using a one-way ANOVA with a post-hoc Tukey HSD test with a p-value threshold equal to 0.01. For each dataset, the technique with the best performance score, and also any other method that is not statistically significantly inferior to it, is bold-faced.

The results indicate that \algo achieves a statistically superior performance (with other methods, in some cases) in 19 out of the 26 datasets. On several datasets, including some of the larger ones, i.e. KDDCup99, and the majority of the MNIST data, BAE shows a very large winning margin across all competing techniques. BAE clearly demonstrates a superior performance against RandNet. The performance of BAE is always superior to that of RandNet, except in four cases (WDBC, MNIST5, MNIST8, and MNIST9), where both BAE and RandNet are in the same statistical significance tier.
 
For Shuttle, Stamps, and Wilt, all methods perform poorly, except HiCS. These results are likely due to the fact that outliers may be hidden in subspaces of the original feature space. 
However, HiCS achieves a superior performance only in four cases, and has
 a poor AUCPR score of $0.5\%$ for KDDCup99, and it does not come even close to BAE in the MNIST scenarios.
 
 Overall BAE is the most robust performer across all datasets, with an average AUCPR score of 46.2\%.  RandNet comes second with 42.6\%, and SAE9 third with 40.0\%. It is interesting to observe how close SAE9 and RandNet are on average. Both single autoencoders (SAE9 and Hawkins) largely outperforms HiCS. HiCS is by far the worst performer on average.


\section{Discussion}
\label{sec: discussion}


In this section we evaluate the impact of the used parameters, the implemented strategies, and discuss some of the analyses that guide our choices. 

\begin{figure*}[h]
    \centering  
	\includegraphics[width=\textwidth]{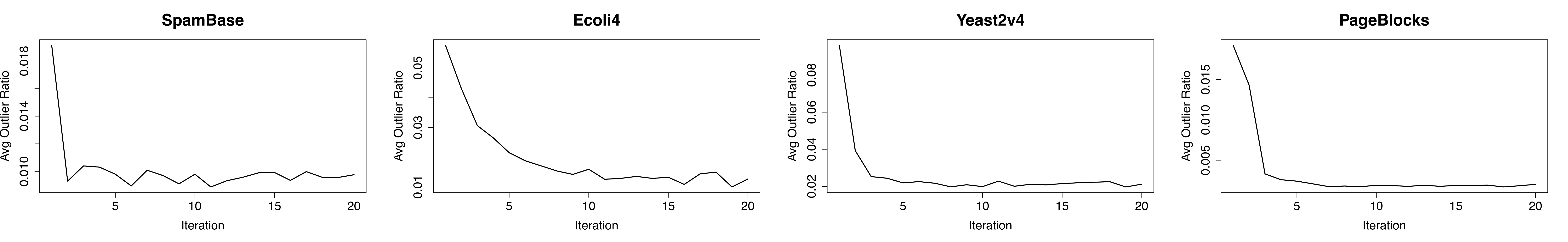}
\caption{Average outlier ratio in the training sample across iterations of \algo for some datasets.}
\label{fig:outlier-ratio}
\end{figure*}

\begin{figure*}[ht]
	\includegraphics[width=\textwidth,scale=0.5]{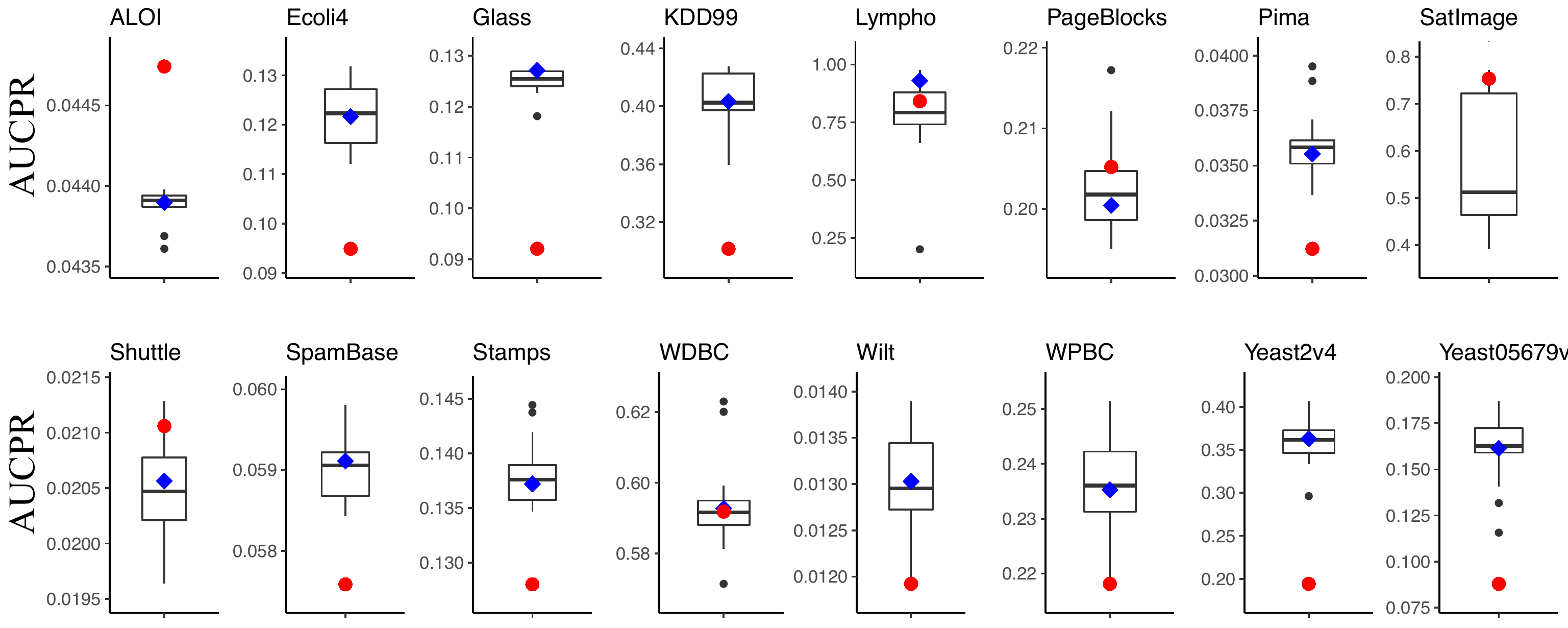}
\caption{Box plots of the ensemble performance of \algo for a singe run. The blue diamond marks the performance of the ensemble consensus, and the red circle marks the performance of SAE9.}
\label{fig:box-plots}
\vspace{-1.0em}
\end{figure*}

\begin{figure*}[ht]
	\includegraphics[width=\textwidth]{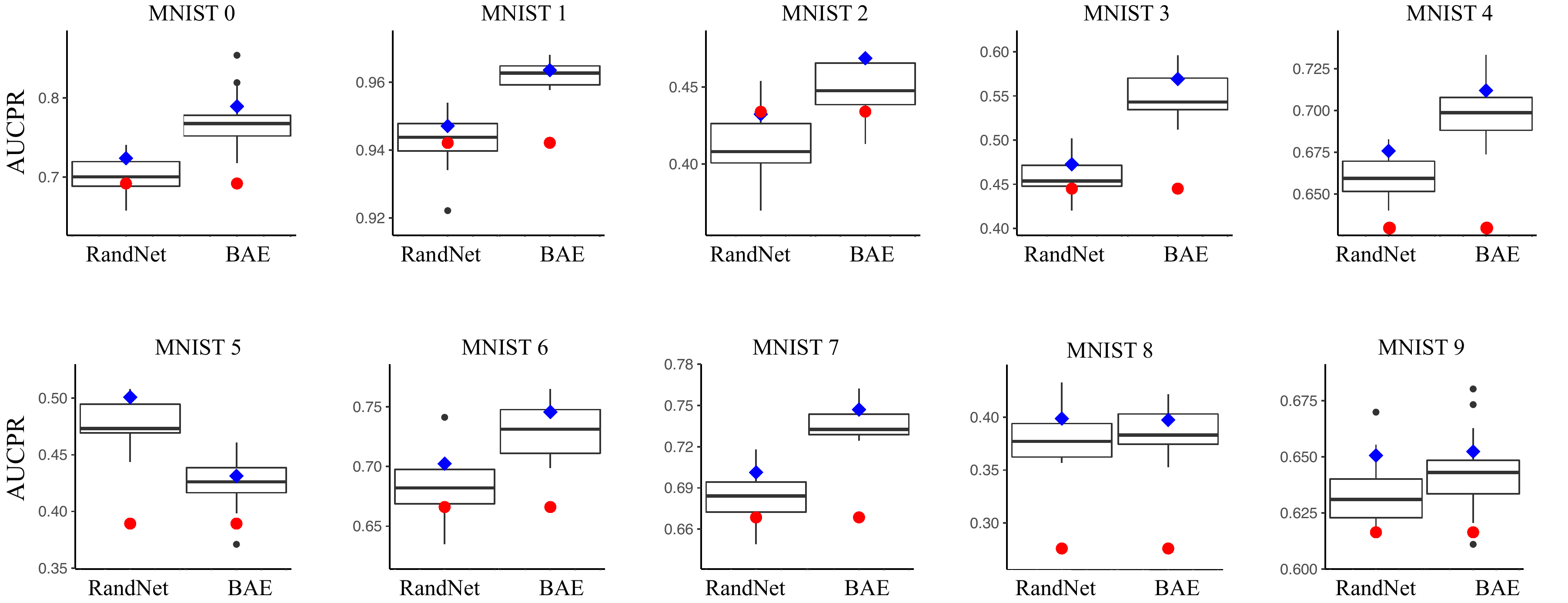}
\caption{Box plots of the ensemble performance of \algo and RandNet on MNIST.}
\label{fig:box-plots-mnist}
\end{figure*}

\input{table_diversity_row.tex}

\textbf{Depth of the \algo components:}\label{subsection:ens-depth}
The performance of \algo is a function of the depth of the autoencoders used in building the ensemble. As discussed in Section~\ref{sec:methodology}, we train \algo on multiple depths of base autoencoders. The optimal depth is then chosen based on the average reconstruction error achieved by each ensemble on their samples. In fact, we choose as optimal depth the one corresponding to the number of layers of the 
ensemble that generates the minimum average reconstruction error.
Figure~\ref{fig:auc_vs_err} demonstrates the inverse relationship between ensemble performance, in AUCPR, and average reconstruction error on training samples. Due to space limitations we show these plots only for some datasets.

\textbf{Outlier ratio variation:}
One of the main objectives of \algo is to reduce the number of outliers in the training samples through a sequence of autoencoders.
In Figure ~\ref{fig:outlier-ratio}, we plot the ratio of outliers belonging to the sampled datasets $\mathbf{X}^{(i)}$ in each iteration $i$ of \algo. In this way, we verify the efficacy of \algo in reducing the overall number of outliers in the different sub-samples. Again, because of spatial limits, we only showcase the decreasing outlier ratio trends for some datasets.






\textbf{Ensemble efficacy:}\label{subsection:ens-efficacy}
In order to study the success of \algo as an ensemble-based method, we look at the box plots of the performance of its components and compare them with those of a single autoencoder, SAE9. In Figure~\ref{fig:box-plots}, the blue diamond shows the performance of \algo and the red circle marks those of SAE9. Note that, in order to show the box plots, we use the scores obtained from a single experiment run. Therefore, the AUCPR values can be different from those reported in Table~\ref{tab:baselines}. We observe that the performance of \algo is either comparable or superior to the median AUCPR. 

\textbf{Diversity induced by the ensemble:}\label{subsec:div}
As discussed above, we exploit ensemble learning in order to create diversity and, consequently, avoid the pitfalls of over-fitting. In order to show the effectiveness of the \algo in injecting diversity, we define a diversity measure for outlier ensembles and compare \algo's scores in different datasets with those of RandNet.

\textit{Diversity measure:} In order to study the diversity that \algo generates, we first need to come up with a diversity measure for an ensemble. Recall that each AE$^{(i)}$ of BAE produces an outlier ranking list denoted as $S^{(i)}$. We define the correlation of an ensemble as the average of the correlation values for every possible pairwise combination of the components. We use Kendall's tau $\kappa^\tau$ as a correlation measure between two rankings $S^{(i)}$ and $S^{(j)}$. Therefore, we compute the diversity of an ensemble as follows:

{$$ D({\{AE\}}_1^{m-1})= 1-2\cdot\frac{  \sum\limits_{i=1}^{m-2}\sum\limits_{j>i}^{m-1} \kappa^\tau\Big({S}^{(i)},{S}^{(j)}\Big)}{(m-1)(m-2)} $$}

Table~\ref{tab:diversity} shows that \algo is  successful in generating more diverse ensembles, except for the MNIST datasets where RandNet achieves higher levels of diversity. However, BAE largely outperforms RandNet on MNIST. To compare the performance of the components of RandNet and \algo,  we compute the corresponding box plots in Figure~\ref{fig:box-plots-mnist}. In all cases, except MNIST5, the components of \algo are more accurate. This analysis corroborates the fact that  RandNet achieves higher diversity at the expense of accuracy. On the contrary, BAE achieves a good trade-off between diversity and accuracy, and thus an overall superior performance.

%% file: tbl/tbl_ds_sizes.tex

\begin{table*}[t]
\centering
\begin{adjustbox}{width=\linewidth}
\begin{tabular}{rcccccccccccccccccccccccccc}
 & \rot{\textbf{ALOI}} & \rot{\textbf{Ecoli4}} & \rot{\textbf{Glass}} & \rot{\textbf{KDDCup99}} & \rot{\textbf{Lympho}} & \rot{\textbf{PageBlocks}} & \rot{\textbf{Pima}} & \rot{\textbf{SatImage}} & \rot{\textbf{Shuttle}} & \rot{\textbf{SpamBase}} & \rot{\textbf{Stamps}} & \rot{\textbf{WDBC}} & \rot{\textbf{Wilt}} & \rot{\textbf{WPBC}} & \rot{\textbf{Yeast05679v4}} & \rot{\textbf{Yeast2v4}} & \rot{\textbf{MNIST0}} & \rot{\textbf{MNIST1}} & \rot{\textbf{MNIST2}} & \rot{\textbf{MNIST3}} & \rot{\textbf{MNIST4}} & \rot{\textbf{MNIST5}} & \rot{\textbf{MNIST6}} & \rot{\textbf{MNIST7}} & \rot{\textbf{MNIST8}} & \rot{\textbf{MNIST9}} \\ \toprule
\multicolumn{1}{l}{\textbf{Instances}} & 49,534 & 336 & 214 & 48,113 & 148 & 4,982 & 510 & 1,105 & 1,013 & 2,579 & 315 & 367 & 4,655 & 198 & 528 & 514 & 7,534 & 8,499 & 7,621 & 7,770 & 7,456 & 6,950 & 7,508 & 7,921 & 7,457 & 7,589 \\
\multicolumn{1}{l}{\textbf{Attributes}} & 27 & 7 & 7 & 40 & 18 & 10 & 8 & 36 & 9 & 57 & 9 & 30 & 5 & 34 & 8 & 8 & 784 & 784 & 784 & 784 & 784 & 784 & 784 & 784 & 784 & 784 \\
\multicolumn{1}{l}{\textbf{Outliers \%}} & 3 & 6 & 4 & 0.4 & 4 & 2 & 2 & 3 & 1 & 2 & 2 & 2.45 & 2 & 2.4 & 10 & 10 & 8 & 7 & 8 & 8 & 8 & 9 & 8 & 8 & 8 & 8 \\ \bottomrule
\end{tabular}
\end{adjustbox}
\caption{Summary of the dataset statistics.}
\label{tab:datasets}
\vspace{-1.0em}
\end{table*}

%% file: Results-table.tex
\begin{table}[ht]
  \centering
      \begin{adjustbox}{width=\columnwidth}
      \setlength\tabcolsep{4.0pt} 
    \begin{tabular}{l c | c | c| c | c | c| c }

Dataset & LOF & SAE9 & OCSVM & Hawkins & HiCS & RandNet & BAE\\
\hline
\textbf{ALOI} & \textbf{13.3} & 4.5 & 4.2 & 4.0 & 3.9 & 4.5 & 4.5  \\
\textbf{Ecoli4} & 5.4 & 8.2 & 12.3 & 12.1 & 4.8 & 9.4 & \textbf{16.0}\\
\textbf{Glass} & \textbf{15.9} & 9.2 & 7.7 & 10.5 & 8.0 & 9.2 & 12.3\\
\textbf{KDDCup99} & 3.1  & 29.2 & 34.8 &  20.1 & 0.5 &  15.8 & \textbf{40.8}\\
\textbf{Lympho} & 48.6 & 74.7 & 79.9 & 23.2 & 4.0 & 85.6 & \textbf{96.7}\\
\textbf{PageBlocks} & 25.2 & 19.6 & 23.8 & \textbf{29.6} & 15.3 & 21.0 & \textbf{31.4}\\
\textbf{Pima} & 1.9 & 3.1 & 2.8 & 3.0 & 3.2 & 3.1 & \textbf{3.5}\\
\textbf{SatImage} & 21.0 & \textbf{63.0} & 52.3 & 29.6 & 2.3 & 57.2 & \textbf{68.0}\\
\textbf{Shuttle} & 11.5 & 2.1 & 2.2 & 3.7 & \textbf{23.7} & 2.1 & 2.1\\
\textbf{SpamBase} & \textbf{7.0} & 5.8 & \textbf{6.6} & 6.0 & 2.0 & 5.8 & 5.9\\
\textbf{Stamps} & 9.9 & 13.5 & 11.7 & 14.2 & \textbf{21.2} & 12.8 & 13.8\\
\textbf{WDBC} & \textbf{54.8} & \textbf{50.7} & \textbf{59.6} & 49.0 & 26.9 & \textbf{59.2} & \textbf{56.7}\\
\textbf{Wilt} & 7.3 & 1.2 & 1.4 & 1.2 & \textbf{16.2} & 1.2 & 1.3\\
\textbf{WPBC} & 21.1 & 21.9 & 22.6 & 21.5 & \textbf{24.9} & 22.1 & 23.9\\
\textbf{Yeast05679v4} & 13.6 & 10.2 & 15.5 & 11.4 & 15.5 & 10.9 & \textbf{18.1}\\
\textbf{Yeast2v4} & 14.8 & 23.5 & \textbf{35.9} & 23.1 & 15.7 & 26.0 & \textbf{37.1}\\
\hline

\textbf{MNIST0} & 18.9 & 66.1 & 49.6 & 46.3 & 19.1 & 71.3 & \textbf{78.9}\\
\textbf{MNIST1} & 14.3 & 92.1 & 90.0 & 93.7 & 9.3 & 94.7 & \textbf{96.1}\\
\textbf{MNIST2} & 17.3 & 42.7 & 27.5 & 37.0 & 13.9 & 43.6 & \textbf{47.9}\\
\textbf{MNIST3} & 19.9 & 45.6 & 37.4 & 45.1 & 18.9 & 47.4 & \textbf{55.4}\\
\textbf{MNIST4} & 24.2 & 57.7 & 46.8 & 51.6 & 15.6 & 67.6 & \textbf{71.2}\\
\textbf{MNIST5} & 28.1 & \textbf{49.7} & 23.3 & \textbf{48.8} & 17.0 & \textbf{49.5} & \textbf{48.4}\\
\textbf{MNIST6} & 23.4 & 67.5 & 42.8 & 53.3 & 18.4 & 70.6 & \textbf{73.8}\\
\textbf{MNIST7} & 18.3 & 66.9 & 52.8 & 59.7 & 17.5 & 69.8 & \textbf{74.8}\\
\textbf{MNIST8} & 17.7 & 32.8 & 27.5 & \textbf{40.3} & 15.4 & \textbf{39.4} & \textbf{39.4}\\
\textbf{MNIST9} & 28.2 & 60.8 & 41.5 & 50.6 & 16.9 & \textbf{64.8} & \textbf{63.9}\\
\hline
\textbf{Average} & 18.5 & 40.0 & 33.0 & 34.8 & 13.5 & 42.6 & \underline{46.2} \\

\hline
\end{tabular}
  \end{adjustbox}
  
  \caption{Average AUCPR values in percentage (over 30 runs) for all methods.
  \label{tab:baselines}}
\end{table}

%% file: table_diversity_row.tex

\begin{table*}[ht]
\centering
\begin{adjustbox}{width=\linewidth}

\begin{tabular}{lccccccccccccc}
 & \textbf{ALOI} & \textbf{Ecoli4} & \textbf{Glass} & \textbf{KDDCup99} & \textbf{Lympho} & \textbf{PageBlocks} & \textbf{Pima} & \textbf{SatImage} & \textbf{Shuttle} & \textbf{SpamBase} & \textbf{Stamps} & \textbf{WDBC} & \textbf{Wilt} \\ \toprule
RandNet & 0.223 & 0.116 & 0.037 & \textbf{0.575} & 0.275 & \textbf{0.383} & 0.048 & \textbf{0.441} & 0.027 & 0.01 & 0.063 & \textbf{0.11} & \textbf{0.143} \\
BAE & \textbf{0.254} & \textbf{0.144} & \textbf{0.095} & 0.283 & \textbf{0.294} & 0.205 & \textbf{0.066} & 0.222 & \textbf{0.085} & \textbf{0.042} & \textbf{0.084} & \textbf{0.095} & \textbf{0.133}\\ \bottomrule  \\

 & \textbf{WPBC} & \textbf{Yeast05679v4} & \textbf{Yeast2v4} & \textbf{MNIST0} & \textbf{MNIST1} & \textbf{MNIST2} & \textbf{MNIST3} & \textbf{MNIST4} & \textbf{MNIST5} & \textbf{MNIST6} & \textbf{MNIST7} & \textbf{MNIST8} & \textbf{MNIST9} \\ \toprule
RandNet & 0.033 & 
\textbf{0.144} & \textbf{0.121} & 0.25 & \textbf{0.284} & \textbf{0.287} & \textbf{0.244} & \textbf{0.236} & \textbf{0.283} & \textbf{0.245} & \textbf{0.214} & \textbf{0.246} & 0.205 \\
BAE & \textbf{0.275} & 0.119 & 0.096 & \textbf{0.263} & 0.252 & 0.245 & 0.228 & \textbf{0.234} & 0.211 & 0.228 & 0.197 & \textbf{0.23} & 0.212\\ \bottomrule
\end{tabular}
\end{adjustbox}
\caption{Diversity of the components generated by RandNet and BAE.}
\label{tab:diversity}
\vspace{-1.0em}
\end{table*}

%% file: sec_conclusion.tex
\section{Conclusion}\label{sec:conclusions}

This paper introduces a boosting-based outlier ensemble approach. \algo trains a sequence of autoencoders by performing a weighted sampling of the data with the aim of progressively reducing the number of outliers in the training. 
This helps to avoid over-fitting, a problem hindering the success of neural networks in anomaly detection.

The progressive reduction of outliers enables the autoencoders to learn better representations of  the inliers, which also results in accurate outlier scores. In addition, each autoencoder is exposed to a different set of outliers, thus promoting diversity among them. Our experimental results show that  BAE achieves  a good accuracy-diversity trade-off, and outperforms state-of-the-art competitors.